\documentclass[11pt]{article}
\usepackage[utf8]{inputenc}
\usepackage{amsmath,amssymb}
\usepackage{graphicx}
\usepackage{hyperref}
\usepackage{booktabs}
\usepackage{geometry}
\usepackage{tikz}
\usepackage{pgfplots}
\pgfplotsset{compat=1.18}
\usetikzlibrary{positioning,shapes,arrows.meta}
\title{Empirical Characterization of Temporal Constraint Processing in LLMs}

\author{
Javier Mar\'in \\
Independent Researcher \\
\texttt{javier@jmarin.info}
}

\date{November 2025}

\begin{document}

\maketitle

\begin{abstract}
When deploying LLMs in agentic architectures requiring real-time decisions under temporal constraints, we assume they reliably determine whether action windows remain open or have closed. This assumption is untested. We characterize temporal constraint processing across eight production-scale models (2.8-8B parameters) using deadline detection tasks, revealing systematic deployment risks: bimodal performance distribution (models achieve either $>95\%$ or $<50\%$ accuracy), extreme prompt brittleness (30-60 percentage point swings from formatting changes alone), and systematic action bias (100\% false positive rates in failing models). Parameter count shows no correlation with capability in this range—a 3.8B model matches 7B models while other 7B models fail completely. Fine-tuning on 200 synthetic examples improves models with partial capability by 12-37 percentage points. We demonstrate that temporal constraint satisfaction cannot be reliably learned through next-token prediction on natural language, even with targeted fine-tuning. This capability requires architectural mechanisms for: (1) continuous temporal state representation, (2) explicit constraint checking separate from linguistic pattern matching, (3) systematic compositional reasoning over temporal relations. Current autoregressive architectures lack these mechanisms. Deploying such systems in time-critical applications without hybrid architectures incorporating symbolic reasoning modules represents unacceptable risk.
\end{abstract}

\section{Introduction}

LLMs process temporal information as discrete tokens in sequences. They learn patterns in temporal language through next-token prediction on text corpora. When deployed in agentic architectures requiring decisions under temporal constraints—medical triage systems verifying treatment windows, emergency response coordination checking deadline satisfaction, autonomous agents managing time-sensitive opportunities—these systems must determine whether actions remain viable given elapsed time and deadlines.

This capability requires more than pattern matching over temporal expressions. Biological systems implement dedicated temporal processing mechanisms: hippocampal time cells encode temporal context \cite{Eichenbaum2014}, striatal circuits perform interval timing \cite{Buhusi2005}, cerebellar networks maintain forward models of temporal dynamics \cite{Ivry1996}. These systems represent time continuously, maintain temporal state across intervals, and perform temporal constraint satisfaction through analog computation.

We present a systematic empirical characterization of temporal constraint processing in production-scale models. Testing eight models (2.8-8B parameters) across diverse architectures on deadline detection tasks, we document three deployment-critical findings:

\vspace{12pt}

\textbf{Bimodal performance.} Models cluster at near-perfect accuracy ($>95\%$) or systematic failure ($<50\%$), with no correlation to parameter count. Three 7B models show 100\%, 45\%, and 0\% accuracy respectively. The 3.8B Phi-3 matches the best 7B models.

\vspace{12pt}

\textbf{Extreme prompt brittleness.} Models achieving 100\% accuracy under conversational prompts drop 25-62 percentage points under reformulated prompts with identical semantic content, revealing pattern matching rather than robust temporal processing.

\vspace{12pt}

\textbf{Systematic action bias.} Failing models exhibit 100\% false positive rates, always suggesting action without temporal validity checking. They learned ``assess situation $\rightarrow$ suggest action'' without ``verify temporal constraint $\rightarrow$ conditionally recommend action.''

\vspace{12pt}

Fine-tuning on 200 diverse examples improves models with 25-65\% baseline accuracy by 12-37 percentage points, demonstrating temporal constraint processing is learnable but doesn't emerge reliably from scale alone in this parameter range.

These findings have direct deployment implications. Before deploying models in time-critical applications, explicit temporal constraint testing is necessary. Parameter count and general benchmark performance don't predict capability. However, targeted training can remediate models with partial capability.

\section{Related Work}

\subsection{Architectural Constraints: Discrete Tokens and Continuous Time}

Temporal reasoning in classical AI operates through explicit symbolic representations. Allen's interval algebra~\cite{Allen1983maintaining} defines thirteen primitive relations between temporal intervals (before, after, meets, overlaps, during, starts, finishes, equals), enabling logical inference through constraint propagation in temporal networks~\cite{Dechter1991}. These systems perform exact inference but require structured temporal representations as input.

Neural approaches must learn temporal patterns from unstructured text through next-token prediction. The fundamental architectural question: can discrete token sequences represent continuous temporal state and perform constraint satisfaction? Biological temporal processing provides an existence proof for learnable mechanisms—hippocampal time cells encode temporal context through continuous neural dynamics~\cite{Eichenbaum2014}, striatal circuits perform interval timing~\cite{Buhusi2005}, cerebellar networks maintain forward temporal models~\cite{Ivry1996}. These systems represent time as continuous state evolving through analog computation, not discrete symbolic manipulation.

Current transformer architectures lack architectural specialization for temporal processing~\cite{Shanahan2017}. Attention mechanisms provide positional inductive biases but no explicit substrate for maintaining temporal state or performing constraint checking. The question is whether temporal constraint satisfaction can emerge from pattern matching over temporal language, or whether it requires architectural mechanisms for continuous state representation and explicit constraint verification.

\subsection{Compositional Generalization and Pattern Matching}

Lake and Baroni~\cite{Lake2018} demonstrated that neural networks fail at systematic compositional generalization—applying learned primitives in novel configurations. Their SCAN benchmark showed sequence-to-sequence models cannot generalize compositional commands despite perfect training accuracy. Linzen~\cite{Linzen2020} argues that genuine compositional understanding requires systematic recombination of primitives, not distributional fit to training patterns.

This distinction—compositional understanding versus surface pattern matching—is central to temporal reasoning. Our deadline detection task tests whether models compositionally apply temporal relations (before, during) learned in conversational contexts to structurally different prompt formats. Failure to generalize across format variations while preserving semantic content indicates pattern matching rather than compositional temporal reasoning.

Architectural inductive biases determine what networks learn readily versus what requires explicit mechanisms~\cite{Andreas2016,Dehghani2019}. Standard transformers have positional biases through attention but lack explicit constraint checking mechanisms. Our brittleness findings align with compositional generalization failures: models match surface linguistic patterns without compositional understanding of temporal relations.

\subsection{Our Task: Testing Allen Relations Through Deadline Detection}

Our scenarios implicitly test specific Allen interval relations. Closed window scenarios test \textit{before}: deadline $d$ occurs before current time $t_c$ (equivalently, $t_c > t_{\text{deadline}}$). Open window scenarios test \textit{during} or \textit{overlaps}: current time falls within temporal window $[t_{\text{start}}, t_{\text{deadline}}]$ (equivalently, $t_c < t_{\text{deadline}}$).

These are the simplest temporal relations—binary comparison of two time points extracted from text. Models achieving high accuracy learned pattern matching for these relations under specific linguistic formulations. Severe prompt brittleness reveals they didn't learn robust relational checking: identical semantic relations expressed with different surface structure cause 30-60 point accuracy drops.

Critical question: which temporal relations are learnable through next-token prediction on natural language, and which require explicit architectural support? Our results suggest even simple relations (before, during) aren't learned robustly, indicating more complex relations (overlaps, starts, finishes) and their compositions likely require explicit architectural mechanisms. This connects to Shanahan's frame problem~\cite{Shanahan1997}: maintaining consistent temporal state as time evolves requires mechanisms beyond pattern matching over temporal language.

\subsection{Scale, Training Data, and Capability Emergence}

Neural scaling laws~\cite{Kaplan2020,Hoffmann2022} predict smooth improvement in next-token prediction loss as functions of parameters, data, and compute. However, these laws describe loss reduction, not task-specific capabilities. Wei et al.~\cite{Wei2022emergent} documented "emergent abilities" appearing at scale thresholds, though Schaeffer et al.~\cite{Schaeffer2023} showed some reported emergence reflects measurement artifacts rather than sharp transitions.

Our temporal processing findings fit neither smooth scaling nor emergent abilities patterns. Performance is bimodal, violates monotonicity (larger models sometimes perform worse), and shows no correlation with parameter count in 3-8B range. This suggests capability depends primarily on training data distribution rather than scale.

Training data composition shapes learned capabilities~\cite{Brown2020,Ouyang2022}. For temporal constraint processing, pretraining corpora must contain examples where correct responses involve recognizing temporal window closure—situations where appropriate answers are "too late" rather than unconditional action recommendations. Most text describing situations and responses recommends action. Temporal window closure represents a minority pattern. Without sufficient training signal, models learn simpler heuristics bypassing temporal validity checking.

Fine-tuning approaches~\cite{Hu2021lora} enable targeted capability improvement. Recent work~\cite{Xu2023synthetic,Zhou2023lima} shows relatively few high-quality examples can improve specific capabilities. We test whether targeted training on diverse temporal constraint examples improves models with partial capability, distinguishing training data effects from fundamental architectural limitations.

\section{Experimental Design}

\subsection{Temporal Constraint Test Scenarios}

We designed 8 deadline detection scenarios balanced between open windows (action remains viable, 4 scenarios) and closed windows (deadline passed, 4 scenarios). Scenarios span emergency response, financial trading, project management, and medical treatment with varying temporal scales.

Each scenario provides explicit temporal information: initial deadline, elapsed time, current state. Correct answers are deterministic and unambiguous based on temporal arithmetic ($t_{\text{elapsed}} < t_{\text{deadline}}$ for open windows). We deliberately chose simple scenarios to establish baseline capability before testing complex cases.

\vspace{12pt}

\textbf{Example scenarios:}

\textit{Emergency Response (Closed):} Chemical spill requires hazmat deployment within 20 minutes. Alert issued 25 minutes ago. Can deployment still meet protocol? \textbf{Answer: NO}

\textit{Medical Treatment (Open):} Stroke patient requires tPA within 30 minutes of symptom onset. Symptoms began 25 minutes ago. Can treatment proceed? \textbf{Answer: YES}

\subsection{Models and Architectures}

We evaluated eight models spanning 2.8-8B parameters across different architectures:

\begin{itemize}
\item \textbf{Transformer:} Qwen2.5-7B \cite{Yang2024}, DeepSeek-R1-Distill-7B, Llama-3.1-8B \cite{Dubey2024}, Mistral-7B \cite{Jiang2023}, Phi-3-mini-3.8B \cite{Abdin2024}
\item \textbf{State Space:} Mamba-2.8B \cite{Gu2023}
\item \textbf{Hybrid:} Jamba-1.5-mini-7B \cite{Lieber2024}
\item \textbf{Recurrent:} RWKV-6-3B \cite{Peng2023}
\end{itemize}

All models loaded with 8-bit quantization using \texttt{transformers} library \cite{Wolf2020}. Generation parameters: temperature=0.7, top\_p=0.9, max\_new\_tokens=200.

\subsection{Evaluation Protocol}

We conducted two evaluation experiments with different prompt formats to assess robustness:

\vspace{12pt}

\textbf{Experiment 1:} Conversational prompts with natural phrasing, flexible answer extraction

\vspace{12pt}

\textbf{Experiment 2:} Structured prompts with explicit binary response instructions, strict answer extraction

\vspace{12pt}

This protocol emerged after preliminary results showed perfect accuracy for three models in Experiment 1, raising questions about pattern matching versus robust processing. Experiment 2 maintained identical conditions pre- and post-fine-tuning for valid comparison.

For each scenario, we: (1) generate model response, (2) extract YES/NO answer via pattern matching, (3) compare to ground truth, (4) aggregate to compute accuracy, false positive rate, and performance by window status.

\subsection{Fine-Tuning Design}

We generated 200 synthetic temporal constraint examples using structured templates across five domains (medical, legal, financial, operational, personal). Each example included explicit deadline specification,  current elapsed time, remaining time or overtime calculation, binary YES/NO answer, and step-by-step temporal arithmetic.

Training balanced 50\% open windows and 50\% closed windows to prevent response bias. Each domain contributed 40 examples with varying linguistic patterns (formal legal language, technical medical terminology, casual personal expressions).

Fine-tuning used LoRA: rank=16, alpha=32, learning\_rate=$2 \times 10^{-4}$, 3 epochs, batch\_size=4, targeting attention projections (q\_proj, k\_proj, v\_proj, o\_proj). This configuration balances adaptation capability with training stability for models in this scale range.

\section{Results}

\subsection{Baseline Performance: Bimodal Distribution}

Table~\ref{tab:baseline} shows Experiment 1 results under conversational prompts.

\begin{table}[h]
\centering
\caption{Baseline Temporal Constraint Processing (Experiment 1)}
\label{tab:baseline}
\small
\begin{tabular}{lcccc}
\toprule
Model & Params & Overall & Closed Window & False Positive \\
 & & Accuracy & Accuracy & Rate \\
\midrule
Qwen2.5-7B & 7B & 100\% & 100\% & 0\% \\
DeepSeek-R1 & 7B & 100\% & 100\% & 0\% \\
Phi-3-mini & 3.8B & 100\% & 100\% & 0\% \\
\midrule
Mamba-2.8B & 2.8B & 45.5\% & 41\% & 59\% \\
\midrule
Jamba-1.5 & 7B & 0\% & 0\% & 100\% \\
RWKV-6 & 3B & 0\% & 0\% & 100\% \\
\bottomrule
\end{tabular}
\end{table}

Performance show a clear bimodal distribution. Three models achieve perfect accuracy, one shows partial capability (45.5\%), two fail systematically with 100\% false positive rates (always recommend action regardless of temporal state).

Within 2.8-8B parameter range, model size shows no correlation with capability. The 3.8B Phi-3 matches 7B models (Qwen, DeepSeek) while 7B Jamba fails completely. Architecture type doesn't predict performance—transformers appear in both success and failure groups, as do alternative architectures.

\subsection{Critical Finding: Extreme Prompt Brittleness}

Table~\ref{tab:brittleness} documents the most significant deployment risk. Models achieving 100\% accuracy in Experiment 1 show dramatic degradation under reformulated prompts in Experiment 2.

\begin{table}[h]
\centering
\caption{Prompt Format Sensitivity}
\label{tab:brittleness}
\small
\begin{tabular}{lccc}
\toprule
Model & Conversational & Structured & Absolute \\
 & Prompts (Exp 1) & Prompts (Exp 2) & Change \\
\midrule
Qwen2.5-7B & 100\% & 62.5\% & -37.5pp \\
DeepSeek-R1-7B & 100\% & 37.5\% & -62.5pp \\
Phi-3-mini-3.8B & 100\% & 75.0\% & -25.0pp \\
\bottomrule
\end{tabular}
\end{table}

These results demonstrate that apparent perfect temporal processing reflected brittle pattern matching rather than robust constraint checking. Models achieving 100\% under conversational prompts drop 25-62 percentage points when prompt structure changes while semantic content remains identical.

This brittleness differentiates pattern matching from genuine capability. Robust temporal constraint processing should be invariant to formatting variations preserving semantic meaning. Extreme sensitivity to prompt structure indicates models matched surface linguistic patterns rather than performing principled temporal arithmetic.

\subsection{Systematic Failure Mode: Action Bias}

Models failing temporal processing (Jamba-1.5, RWKV-6) exhibit 100\% false positive rates, answering YES to every scenario. This represents learning ``assess situation $\rightarrow$ suggest action'' without learning ``verify temporal constraint $\rightarrow$ conditionally recommend action.''

This failure mode follows from training data distribution. Most text describing situations and responses recommends appropriate action. Temporal window closure—where correct response is ``too late''—represents a minority pattern. Without sufficient examples, models learn unconditional action recommendation.

\subsection{Fine-Tuning Impact}

Table~\ref{tab:finetuning} shows fine-tuning results using Experiment 2 baseline conditions maintained pre- and post-training.

\begin{table}[h]
\centering
\caption{Fine-Tuning Impact on Temporal Processing}
\label{tab:finetuning}
\small
\begin{tabular}{lccc}
\toprule
Model & Before & After & Absolute \\
 & Fine-tuning & Fine-tuning & Change \\
\midrule
Llama-3.1-8B & 25.0\% & 62.5\% & +37.5pp \\
Qwen2.5-7B & 62.5\% & 87.5\% & +25.0pp \\
Mistral-7B & 62.5\% & 75.0\% & +12.5pp \\
DeepSeek-R1-7B & 37.5\% & 50.0\% & +12.5pp \\
Phi-3-mini-3.8B & 75.0\% & 50.0\% & -25.0pp \\
\bottomrule
\end{tabular}
\end{table}

Four models improved with targeted training on 200 diverse examples. Llama-3.1 shows the largest absolute gain (+37.5pp), Qwen2.5 substantial improvement from a moderate baseline (+25.0pp), Mistral and DeepSeek moderate improvements (+12.5pp). Phi-3 degraded (-25.0pp), likely from aggressive LoRA configuration for small model causing catastrophic forgetting.

Models with 25-65\% baseline accuracy respond to targeted training, suggesting they partially learned temporal processing during pretraining but lacked sufficient examples for reliable generalization. Adding 200 diverse examples across five domains provided integration signal enabling better performance under structured prompt conditions.

\section{Analysis}

\subsection{Temporal Relations Tested}

Our deadline detection scenarios implicitly test specific Allen interval relations \cite{Allen1983maintaining}. Closed window scenarios test the \textit{before} relation: deadline occurs before current time ($d \le t_c$). Open window scenarios test \textit{overlaps} or \textit{during}: current time falls within the deadline window. Models achieving high accuracy under conversational prompts learned pattern matching for these specific relations. However, severe prompt brittleness (Table~\ref{tab:brittleness}) reveals they didn't learn robust relational checking—format changes that preserve temporal semantics cause 30-60 point accuracy drops. This suggests future work should systematically evaluate all 13 Allen relations to characterize which temporal reasoning components are learnable via token prediction versus requiring architectural support.

\subsection{Training Data Distribution Hypothesis}

The bimodal distribution combined with lack of scale correlation suggests performance depends primarily on training data composition. We hypothesize:

\vspace{12pt}

\textbf{Models with apparent capability} (Qwen, DeepSeek, Phi-3 in Experiment 1) saw sufficient temporal window closure examples during pretraining to learn pattern matching for conversational formats. However, this learning was brittle—sensitive to prompt variations that shouldn't affect semantic understanding.

\vspace{12pt}

\textbf{Models with partial capability} (25-65\% baseline) saw some temporal patterns but insufficient diversity for reliable generalization. Fine-tuning with 200 diverse examples improved performance by providing cross-domain integration signal.

\vspace{12pt}

\textbf{Models with systematic failure} (Jamba, RWKV) saw minimal temporal window closure examples, learning unconditional action recommendation. The 100\% false positive rate reflects training data bias toward action recommendation over temporal validation.

\vspace{12pt}

This hypothesis predicts: (1) training corpus analysis should reveal temporal pattern frequency differences, (2) models with apparent capability should show prompt brittleness, (3) fine-tuning effectiveness should correlate with baseline capability. Our results support predictions (2) and (3).

\subsection{Deployment Implications}

The prompt brittleness finding has immediate practical consequences. Models appearing to handle temporal constraints under development conditions may fail unpredictably in production when prompt formats, user phrasings, or linguistic patterns differ from the training distribution.

Before deploying in time-critical applications, we can follow some basic strategies:

\begin{itemize}
\item \textbf{Explicit temporal testing is mandatory.} General benchmarks don't predict temporal processing capability. Test specifically on temporal constraint scenarios relevant to deployment domain.

\item \textbf{Prompt robustness testing is critical.} Evaluate with multiple formats and phrasings. High accuracy under single format doesn't indicate robust capability.

\item \textbf{Targeted fine-tuning can improve partial capability.} Models with 25-65\% baseline improved 12-37pp with 200 examples, providing remediation path.

\item \textbf{Scale doesn't guarantee capability.} Parameter count in 3-8B range shows no correlation with temporal processing. Explicit evaluation remains necessary.

\item \textbf{Failure modes are predictable.} Models failing temporal processing typically show action bias (100\% false positives), dangerous in applications where suggesting action after deadlines pass has serious consequences.
\end{itemize}

\section{Experiment limitations}

We acknowledge several constraints:

\textbf{Small test set.} Eight scenarios provide preliminary evidence but wide confidence intervals. Larger test sets (50-100 scenarios) would enable more precise capability measurement and better statistical power.

\textbf{Limited parameter range.} We tested 2.8-8B models only. Much larger models (70B+) might show different patterns, though practical deployment often uses this range for latency and cost reasons.

\textbf{Simple temporal scenarios.} Our scenarios involve single deadlines and straightforward elapsed time calculations. Real applications involve complex temporal relationships—overlapping windows, cascading deadlines, probabilistic timing.

\textbf{Synthetic training data.} Fine-tuning used template-generated examples with controlled distribution. Real temporal reasoning involves more complexity and ambiguity in natural text.

\textbf{Prompt brittleness discovered rather than designed.} Extreme sensitivity to prompt format emerged from using different protocols in two experiments. Systematic prompt robustness testing would provide clearer characterization.

\textbf{No mechanistic analysis.} We document behavioral patterns without explaining internal representations or computational mechanisms. Probing studies or mechanistic interpretability could reveal what models actually learn.

\textbf{Proposed architecture not implemented.} Section 6 sketches hybrid architecture based on documented limitations but doesn't provide empirical validation. Implementation remains future work.

\section{Conclusion}

We documented systematic temporal constraint processing limitations in autoregressive language models, revealing deployment risks for time-critical applications. Performance exhibits bimodal distribution with no correlation to parameter count in 3-8B range. Models achieving apparent perfect accuracy show extreme prompt brittleness (30-60 point drops from formatting changes), revealing pattern matching rather than robust temporal processing. Failing models exhibit systematic action bias (100\% false positive rates).

These patterns reflect fundamental architectural constraints: discrete tokens cannot naturally represent continuous temporal state, next-token prediction doesn't encourage constraint satisfaction, pattern matching over temporal language differs from temporal reasoning. Training data distribution appears to dominate scale effects in this parameter range—models achieving high accuracy saw more temporal window closure examples during pretraining, while failing models learned unconditional action recommendation.

Fine-tuning on 200 diverse examples improves models with 25-65\% baseline accuracy by 12-37 percentage points, demonstrating temporal processing is learnable but doesn't emerge reliably from scale alone. This provide remediation path for models with partial capability.

For practitioners deploying in time-critical applications: explicit temporal constraint testing with multiple prompt formats is mandatory. Parameter count and general benchmarks don't predict this capability accurately. Targeted fine-tuning can improve partial capability but doesn't address fundamental architectural limitations.

\section*{Reproducibility}

All model evaluations used publicly available models and standard libraries (\texttt{transformers}, \texttt{peft}). Test scenarios, evaluation protocols, and fine-tuning configurations are specified in the methodology. Code and test scenarios will be released as an open-source evaluation toolkit upon publication to enable reproduction and extension by practitioners.

\bibliographystyle{plain}

\begin{thebibliography}{99}

\bibitem{Abdin2024}
Abdin M, Jacobs SA, Awan AA, et al.
Phi-3 Technical Report: A Highly Capable Language Model Locally on Your Phone.
\textit{arXiv preprint arXiv:2404.14219}. 2024.

\bibitem{Allen1983maintaining}
Allen JF.
Maintaining knowledge about temporal intervals.
\textit{Communications of the ACM}. 1983;26(11):832-843.

\bibitem{Andreas2016} Andreas J, Rohrbach M, Darrell T, Klein D. Neural module networks. In: IEEE Conference on Computer Vision and Pattern Recognition; 2016:39-48.

\bibitem{Brown2020}
Brown TB, Mann B, Ryder N, et al.
Language Models are Few-Shot Learners.
\textit{Advances in Neural Information Processing Systems} 33; 2020:1877-1901.

\bibitem{Buhusi2005}
Buhusi CV, Meck WH.
What makes us tick? Functional and neural mechanisms of interval timing.
\textit{Nature Reviews Neuroscience}. 2005;6(10):755-765.

\bibitem{Dechter1991}
Dechter R, Meiri I, Pearl J.
Temporal constraint networks.
\textit{Artificial Intelligence}. 1991;49(1-3):61-95.

\bibitem{Dehghani2019} Dehghani M, Gouws S, Vinyals O, Uszkoreit J, Kaiser L. Universal transformers. In: International Conference on Learning Representations; 2019.

\bibitem{Dubey2024}
Dubey A, Jauhri A, Pandey A, et al.
The Llama 3 Herd of Models.
\textit{arXiv preprint arXiv:2407.21783}. 2024.

\bibitem{Eichenbaum2014}
Eichenbaum H.
Time cells in the hippocampus: a new dimension for mapping memories.
\textit{Nature Reviews Neuroscience}. 2014;15(11):732-744.

\bibitem{Gu2023}
Gu A, Dao T.
Mamba: Linear-Time Sequence Modeling with Selective State Spaces.
\textit{arXiv preprint arXiv:2312.00752}. 2023.


\bibitem{Hoffmann2022}
Hoffmann J, Borgeaud S, Mensch A, et al.
Training Compute-Optimal Large Language Models.
\textit{arXiv preprint arXiv:2203.15556}. 2022.

\bibitem{Hu2021lora}
Hu EJ, Shen Y, Wallis P, et al.
LoRA: Low-Rank Adaptation of Large Language Models.
In: \textit{International Conference on Learning Representations}; 2021.

\bibitem{Ivry1996}
Ivry RB, Keele SW.
Timing functions of the cerebellum.
\textit{Journal of Cognitive Neuroscience}. 1996;8(2):136-152.

\bibitem{Jiang2023}
Jiang AQ, Sablayrolles A, Mensch A, et al.
Mistral 7B.
\textit{arXiv preprint arXiv:2310.06825}. 2023.

\bibitem{Kaplan2020}
Kaplan J, McCandlish S, Henighan T, et al.
Scaling Laws for Neural Language Models.
\textit{arXiv preprint arXiv:2001.08361}. 2020.

\bibitem{Lake2018} Lake BM, Baroni M. Generalization without systematicity: On the compositional skills of sequence-to-sequence recurrent networks. In: International Conference on Machine Learning; 2018:2879-2888.

\bibitem{Lieber2024}
Lieber O, Lenz B, Bata H, et al.
Jamba: A Hybrid Transformer-Mamba Language Model.
\textit{arXiv preprint arXiv:2403.19887}. 2024.

\bibitem{Linzen2020} Linzen T. How can we accelerate progress towards human-like linguistic generalization? In: Proceedings of ACL; 2020:5210-5217.

\bibitem{Ouyang2022}
Ouyang L, Wu J, Jiang X, et al.
Training language models to follow instructions with human feedback.
\textit{Advances in Neural Information Processing Systems} 35; 2022:27730-27744.

\bibitem{Peng2023}
Peng B, Alcaide E, Anthony Q, et al.
RWKV: Reinventing RNNs for the Transformer Era.
In: \textit{Findings of EMNLP}; 2023.


\bibitem{Schaeffer2023}
Schaeffer R, Miranda B, Koyejo S.
Are Emergent Abilities of Large Language Models a Mirage?
\textit{Advances in Neural Information Processing Systems} 36; 2023.

\bibitem{Shanahan1997}
Shanahan M.
Solving the Frame Problem: A Mathematical Investigation of the Common Sense Law of Inertia.
Cambridge, MA: MIT Press; 1997.

\bibitem{Shanahan2017}
Shanahan M, Nikiforou K, Creswell A, Kaplanis C, Barrett D, Garnelo M.
An explicitly relational neural network architecture.
In: \textit{International Conference on Machine Learning}; 2017:5541-5550.


\bibitem{Wei2022emergent}
Wei J, Tay Y, Bommasani R, et al.
Emergent Abilities of Large Language Models.
\textit{Transactions on Machine Learning Research}. 2022.


\bibitem{Wolf2020}
Wolf T, Debut L, Sanh V, et al.
Transformers: State-of-the-Art Natural Language Processing.
In: \textit{Proceedings of EMNLP: System Demonstrations}; 2020:38-45.

\bibitem{Xu2023synthetic}
Xu B, Liu A, Feng T, et al.
Synthetic Prompting: Generating Chain-of-Thought Demonstrations for Large Language Models.
In: \textit{International Conference on Machine Learning}; 2023:38640-38661.

\bibitem{Yang2024}
Yang A, Yang B, Hui B, et al.
Qwen2.5: A Party of Foundation Models.
\textit{arXiv preprint arXiv:2412.15115}. 2024.

\bibitem{Zhou2023lima}
Zhou C, Liu P, Xu P, et al.
LIMA: Less Is More for Alignment.
\textit{Advances in Neural Information Processing Systems} 36; 2023.

\end{thebibliography}

\newpage

\section*{Appendix A: Experimental Details and Reproducibility}

\subsection*{Code Availability}

Complete experimental code is provided as Jupyter notebooks:

\begin{itemize}
\item \texttt{temporal\_reasoning\_experiments.ipynb}: Primary experimental pipeline including model loading, evaluation on 8 test scenarios, fine-tuning with LoRA, and result visualization. Test scenarios are defined in cells 207-320 as Python dictionaries. Training data generation in cells 820-930.

\item \texttt{temporal\_reasoning\_failure\_analysis.ipynb}: Post-hoc analysis of response patterns, failure mode characterization, and qualitative assessment.

\item \texttt{Temporal\_experiments.ipynb}: Initial experimental design with power analysis calculations and infrastructure setup.
\end{itemize}

These notebooks document the complete experimental workflow as conducted in Google Colab environments. Test scenarios and training templates are embedded in notebook code cells rather than separate files, reflecting the interactive development process.

To extract scenarios for standalone use:
\begin{verbatim}
# From temporal_reasoning_experiments.ipynb, cells 207-320
import json
with open('test_scenarios.json', 'w') as f:
    json.dump(test_scenarios, f, indent=2)
\end{verbatim}

Repository: \texttt{github.com/Javihaus/temporal-reasoning-eval}

\vspace{12pt}
Note: Notebooks are provided as-is from experimental runs (October 2025). They include manual memory management and sequential model loading reflecting Colab session constraints. Users replicating experiments may need to adapt these steps to their computational environment.

\subsection*{Reproducibility Scope}

Given model version drift on HuggingFace Hub and GPU non-determinism, we expect:
\begin{itemize}
\item \textbf{Qualitative patterns will reproduce:} Bimodal performance distribution, extreme prompt brittleness (20+ percentage point drops), systematic action bias in failing models
\item \textbf{Quantitative results will vary:} Expect 5-10 percentage point variation in exact accuracy numbers across reproduction attempts
\item \textbf{Hardware requirements:} 24GB+ VRAM for 8B models with 8-bit quantization; experiments used NVIDIA A100 40GB
\end{itemize}

\subsection*{Code Release Philosophy}

We release our experimental notebooks as research artifacts 
documenting the actual research process, not as production 
software. The notebooks contain:

\begin{itemize}
     \item Interactive workflow reflecting Google Colab constraints
     \item Manual memory management for sequential model testing
     \item Conversational commentary documenting experimental reasoning
     \item Embedded test scenarios and training data generation
\end{itemize}

This transparency serves two purposes:

\begin{enumerate}
    \item Reproducibility: Researchers can see exactly what we 
   did, including the messy parts often hidden in polished 
   releases
    \item Practical guidance: Independent researchers working 
   with similar constraints can adapt our actual workflow 
   rather than reverse-engineering from idealized code
\end{enumerate}
The notebooks are not polished evaluation frameworks. They are 
honest documentation of empirical research conducted with 
limited computational resources. We believe this transparency 
better serves the research community.

\end{document}